\documentclass[final]{cvpr}
\newcommand{\final}{0}
\usepackage{xcolor}
\usepackage{times}
\usepackage{epsfig}
\usepackage{graphicx}
\usepackage{amsmath}
\usepackage{amssymb}
\usepackage{multirow}
\usepackage{booktabs}
\usepackage{subfigure}
\usepackage{microtype}
\usepackage{algorithm}  
\usepackage{algpseudocode}  
\usepackage{engord}
\usepackage{ifthen}
\definecolor{WeimingColor}{rgb}{0,0,0.8}
\definecolor{XingjiaColor}{rgb}{0.0,0.1,0.9}
\definecolor{FanColor}{rgb}{0.8,0,0.8}
\definecolor{SaulColor}{rgb}{0.8,0.1,0}
\definecolor{YingguoColor}{rgb}{0.0,0.8,0.4}
\definecolor{XavierColor}{rgb}{0.8,0,0.8}

\newcommand{\weiming}[1]{{\color{WeimingColor} [Weiming: #1]}}
\newcommand{\xingjia}[1]{{\color{XingjiaColor}[Xingjia: #1]}}
\newcommand{\saul}[1]{{\color{SaulColor}[Saul: #1]}}
\newcommand{\xavier}[1]{{\color{XavierColor}[Xavier: #1]}}
\newcommand{\fan}[1]{{\color{FanColor}[Fan: #1]}}
\newcommand{\yingguo}[1]{{\color{YingguoColor}[Yingguo: #1]}}

\newcommand{\warning}[1]{{\it\color{red} #1}}
\newcommand{\toremove}[1]{{\it\color{red} (To remove) #1}}
\newcommand{\note}[1]{{\it\color{blue} #1}}
\newcommand{\nothing}[1]{}

\ifthenelse{\equal{\final}{1}}
{
\renewcommand{\weiming}[1]{}
\renewcommand{\fan}[1]{}
\renewcommand{\xingjia}[1]{}
\renewcommand{\saul}[1]{}
\renewcommand{\xavier}[1]{}
\renewcommand{\yingguo}[1]{}

\renewcommand{\warning}[1]{}
\renewcommand{\toremove}[1]{}
\renewcommand{\note}[1]{}
\renewcommand{\nothing}[1]{}
}{}

\usepackage[pagebackref=true,breaklinks=true,colorlinks,bookmarks=false]{hyperref}

\def\cvprPaperID{7272} 
\def\confYear{CVPR 2021}

\begin{document}

\title{Unveiling the Potential of Structure Preserving for \\ Weakly Supervised Object Localization}

\author{Xingjia Pan$^{1,3 \ast}$\quad Yingguo Gao$^{1 \ast}$\quad Zhiwen Lin$^{1}$\thanks{Equal contribution}\quad Fan Tang$^{2}$\thanks{Corresponding author} \quad Weiming Dong$^{3,4,5}$\quad \\ Haolei Yuan$^{1}$\quad Feiyue Huang$^{1}$ \quad Changsheng Xu$^{3,4,5}$\\
$^1$Youtu Lab, Tencent \quad $^2$Jilin University \quad $^3$NLPR, Institute of Automation, CAS\quad \\$^4$School of Artificial Intelligence, UCAS \quad $^5$CASIA-LLVision Joint Lab\\
{\tt\small \{noahpan, yingguogao, xavierzwlin,harryyuan,garyhuang\}@tencent.com} \\
{\tt\small tangfan@jlu.edu.cn}, {\tt\small \{weiming.dong, changsheng.xu\}@ia.ac.cn}

}
\maketitle

\begin{abstract}
Weakly supervised object localization (\textit{WSOL}) remains an open problem given the deficiency of finding object extent information using a classification network. 
Although prior works struggled to localize objects through various spatial regularization strategies, we argue that how to extract object structural information from the trained classification network is neglected. 
In this paper, we propose a two-stage approach, termed structure-preserving activation (SPA), toward fully leveraging the structure information incorporated in convolutional features for WSOL.
%
First, a restricted activation module (RAM) is designed to alleviate the structure-missing issue caused by the classification network on the basis of the observation that the unbounded classification map and global average pooling layer drive the network to focus only on object parts.
Second, we designed a post-process approach, termed self-correlation map generating (SCG) module to obtain structure-preserving localization maps on the basis of the activation maps acquired from the first stage. 
Specifically, we utilize the high-order self-correlation (HSC) to extract the inherent structural information retained in the learned model and then aggregate HSC of multiple points for precise object localization. 
Extensive experiments on two publicly available benchmarks including CUB-200-2011 and ILSVRC show that the proposed SPA achieves substantial and consistent performance gains compared with baseline approaches.
Code and models are available at \href{https://github.com/Panxjia/SPA_CVPR2021}{\color{magenta}github.com/Panxjia/SPA\_CVPR2021}.
\end{abstract}

\section{Introduction}
\label{sec:intro}
Weakly supervised object localization (WSOL) requires the image-level annotations indicating the presence or absence of a class of objects in images to learn localization models~\cite{papandreou2015weakly,pathak2015constrained,roy2017combining,xue2019danet,wan2018min,wan2019c}. 
In recent years, WSOL has attracted increasing attention because it can leverage rich Web images with tags to learn object-level models.

\begin{figure}[!tb]
\centering
\subfigure{
\includegraphics[width=0.23\linewidth]{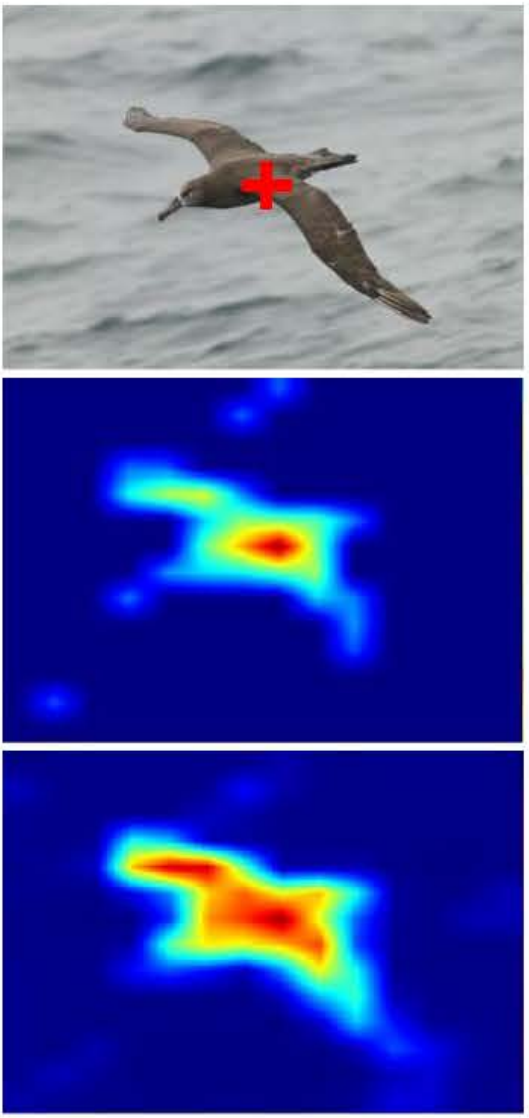}}
\subfigure{
\includegraphics[width=0.215\linewidth]{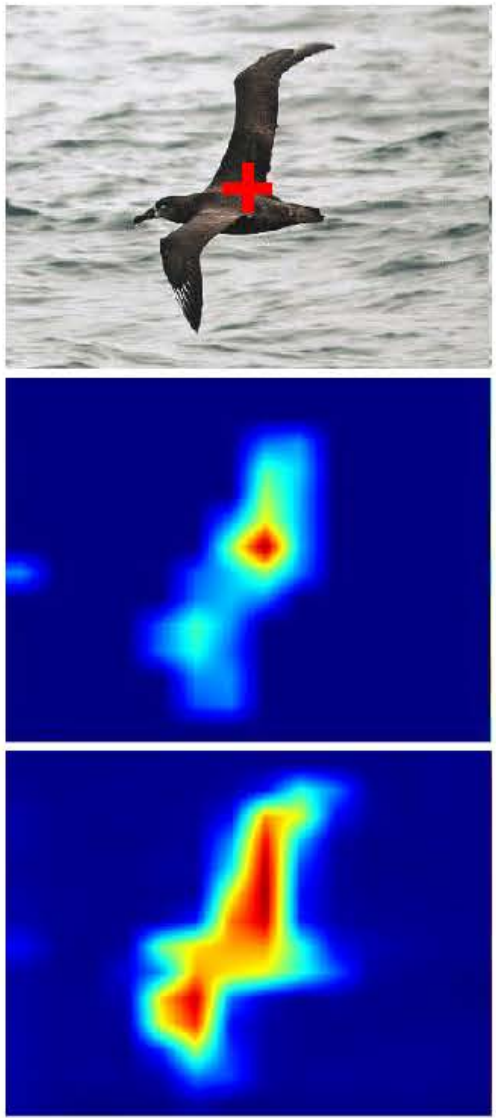}}
\subfigure{
\includegraphics[width=0.24\linewidth]{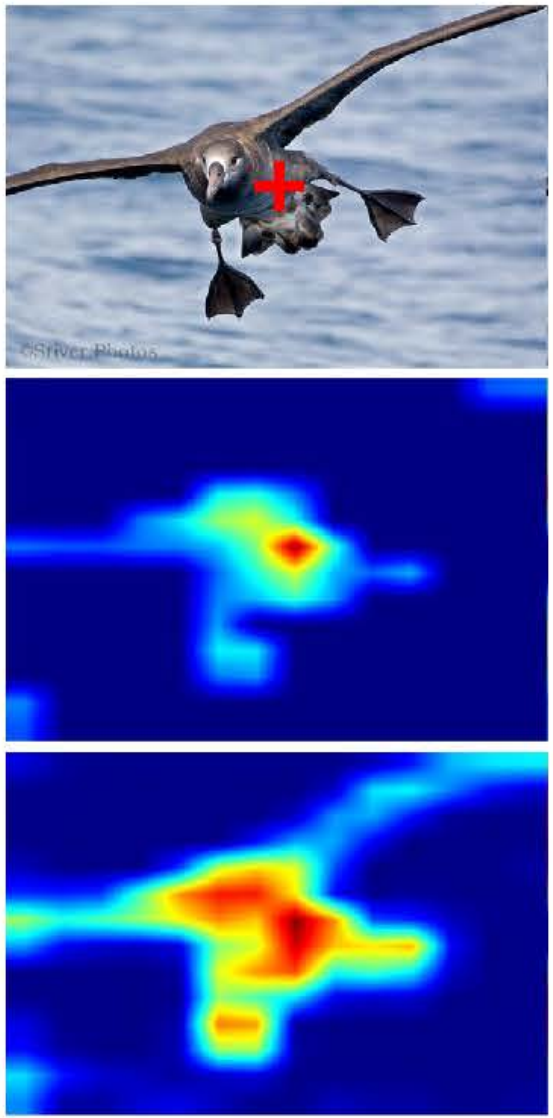}}
\subfigure{
\includegraphics[width=0.216\linewidth]{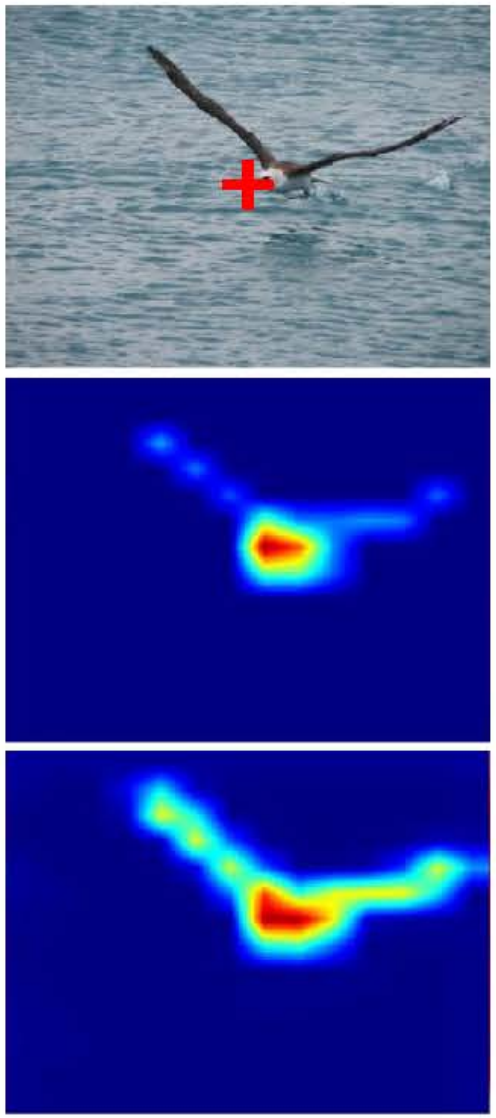}}
\caption{Self-correlation maps corresponding to the image positions masked by red crosses. 
The \engordnumber{2} and \engordnumber{3} rows are the first- and second-order self-correlation maps, respectively.}
\label{fig:hsc_intro}
\end{figure}
As a work for WSOL, Class Activation Mapping (CAM)~\cite{zhou2016learning} uses the intermediate classifier activation to discover discriminative image regions for target object localization~\cite{choe2020evaluating}. 
%
%
Afterward, divergent activation methods~\cite{singh2017hide,xue2019danet,yun2019cutmix} design multiple parallel branches or introduce attention modules to drive networks learning complete object extent. 
%
%
Adversarial erasing methods~\cite{choe2019attention,mai2020erasing,wei2017object,zhang2018adversarial}  pursue learning full object extent in a hide-and-seek fashion.
%
%
Existing methods~~\cite{mai2020erasing,wei2017object,zhang2018adversarial} largely depend on CAM and spatial regularization to localize objects, $e.g.$, expanding activated regions to full object extent; however, preserving the structure of object is unfortunately neglected.

Through experiments, we observed that the head structure of the classification network causes the CAM missing object structure information contained in convolutional features.
Specifically, the network driven by sole classification loss tends to activate a small proportion of features with largest discriminative capability while depressing the majority of object extent. 
Furthermore, many CAM-based methods employ a \emph{global average pooling} (GAP)~\cite{lin2013network} layer atop the feature maps to retain the localization information.
The GAP layer treats each pixel within the feature map equally, which hinders distinguishing true objects from noisy background.

The inherent spatial correlation in CNNs has been widely used in image classification and object detection areas~\cite{Cao_2019_ICCV,Wang_2018_CVPR,wang2020self}, but remains unexploited for WSOL.
%
Following the self-attention mechanism, recent works~\cite{wang2020self,zhang2020inter,zhang2020rethinking} adopted the first-order pixel-wise correlation to improve object localization. 
However, pixels of one object usually have dissimilar features for the large appearance variation, which limits the capability of the first-order self-correlation to preserve object structural information, Fig.~\ref{fig:hsc_intro} (\engordnumber{2} row).
%

In this study, we propose a two-stage WSOL approach, termed structure-preserving activation (SPA), for accurate object localization using sole image-level category labels as supervision.
%
First, a restricted activation module (RAM) is designed to avoid the misleading by local extremely high response for classification by suppressing the response value range of CAM and differentiate objects from background under the guidance of estimated pseudo-masks. 
Second, a self-correlation map generating (SCG) module is proposed to refine the localization map under the guidance of the structural information extracted from trained features. 
In SCG, to guide the activation of objects, we propose to use the high-order self-correlation ($HSC$) which facilities capturing precise spatial layouts of objects by long-range spatial correlations, Fig.~\ref{fig:hsc_intro} (\engordnumber{3} row).
%
%
We conduct extensive experiments on the CUB-200-2011~\cite{wah2011caltech} and ILSVRC~\cite{russakovsky2015imagenet}. 
Our method obtains significant gains compared with baseline methods and achieve comparable results with the SOTAs on bounding box and mask localization.

The contributions of this study include:
\begin{itemize}
    \item We unveil that spatial structure preserving is crucial to discover the localization information contained in convolutional features for WSOL. 
    \item We propose a simple-yet-effective SPA approach to distill the structure-preserving ability of features for accurate object localization.
    \item With negligible computational complexity and cost overheads, our proposed approach shows consistent and substantial gains across CUB-200-2011 and ILSVRC datasets for bounding box and mask localization.
\end{itemize}
\section{Related Work}
\label{sec:rw}

\noindent \textbf{Weakly supervised object localization (WSOL)} aims to learn the localization of objects with only image-level labels.
A representative work on WSOL is CAM~\cite{zhou2016learning}, which produces localization maps by aggregating deep feature maps using a class-specific fully connected layer. 
Hwang and Kim~\cite{Hwang2016selftransfer} simplified CAM by removing the last fully connected layer.
Although CAM-based methods are simple and effective, they only identify small discriminative part of objects. 
To improve the activation of CAMs, HaS~\cite{singh2017hide} and CutMix~\cite{singh2017hide} adopted an erasing-based strategy from input images to force the network to focus on more relevant parts of objects.
Differently, ACoL~\cite{zhang2018adversarial} and ADL~\cite{choe2019attention} instead erased feature maps corresponding to discriminative regions and used multiple parallel classifiers that were trained adversarially. 
Apart from the above erasing methods, SPG~\cite{zhang2018self} and I$^{2}$C~\cite{zhang2020inter} increased the quality of localization maps by introducing the constraint of pixel-level correlations into the network. 
DANet~\cite{xue2019danet} applied a divergent activation to learn complementary and discriminative visual patterns for WSOL.
SEM~\cite{zhang2020rethinking} refined the localization maps by using the point-wise similarity within seed regions. 
GC-Net \cite{lu2020geometry} took geometric shape into account and proposed a multi-task loss function.
Given that existing methods only focus on expanding activation regions, they are challenged by the contradiction between precise classification and object localization. The problem about how to leverage a classification network to active and localize full object extent remains unsolved.

\noindent\textbf{Weakly supervised semantic segmentation (WSSS)} aims to predict precise pixel-level object masks using weak annotations.
The mainstream methods for WSSS with image-level labels train classification networks to estimate object localization maps as pseudo masks which are further used for training the segmentation networks. 
To generate accurate pseudo masks, \cite{kolesnikov2016seed,Ahn_2018_CVPR,Huang_2018_CVPR,Wang_2018_CVPR} resorted to region growing strategy. 
Meanwhile, some researchers investigated to directly enhance the feature-level activated regions~\cite{Lee_2019_CVPR,Wei_2018_CVPR,zhang2018interpretable}.
Others accumulated CAMs through multiple training phases~\cite{Jiang_2019_ICCV}, exploring boundary constraint~\cite{Chen_2020_ECCV}, leveraging equivariance for semantic segmentation~\cite{wang2020self}, and mining cross-image semantics~\cite{Sun_2020_ECCV} to obtain more perfect pseudo masks.
Recently, researchers found saliency maps offer higher quality heuristic cues than attention maps~\cite{fan2020s4net}.  
\begin{figure*}[!t]
\centering
\includegraphics[width=0.9\linewidth]{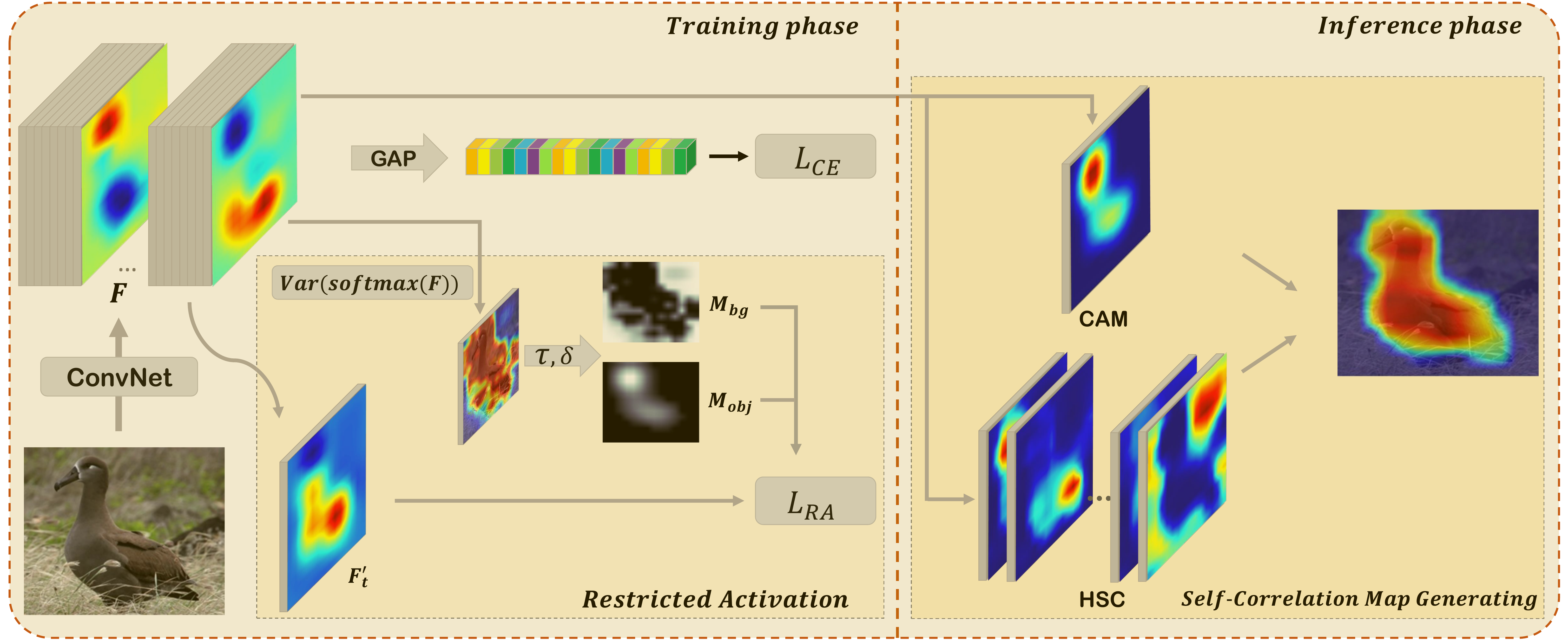}
\caption{Framework of the proposed SPA approach. During the training phase, we designed a restricted activation module paralleled with the classification branch on the baseline architecture of simplified CAM~\cite{zhou2016learning}. In the inference phase, we proposed a self-correlation map generating module, which is a post-process method to refine the localization map by introducing the structural information of objects.}
\label{fig:overview}
\end{figure*}

\noindent\textbf{Feature Self-Correlation.} 
Spatial self-correlation is an instantiate of self-attention mechanism for non-sequential data in computer vision.
Most WSOL/WSSS methods~\cite{wang2020self,zhang2020inter,zhang2020rethinking} utilize the similarity of pixels to refine the features or activation maps following self-attention mechanism.
Wang \etal~\cite{Wang_2018_CVPR} proposed a non-local block to capture long-range dependency within image pixels. 
Cao \etal~\cite{Cao_2019_ICCV} found that the global contexts modeled by non-local network are almost the same for query positions and thereby proposed NLNet~\cite{Wang_2018_CVPR} with SENet~\cite{Hu_2018_CVPR} for global context modeling. 
MST~\cite{song_2019_nips_learnable} proposed the learnable tree filter to leverage the structural property of minimal spanning tree to model long-range dependencies. 
DNL~\cite{yin2020disentangled} disentangled the non-local block into a whitened pairwise term and a unary term to facilitate the learning process.
These methods belong to the first-order self-correlation, and acquire the long-range context by stacking numerous modules in different stages. 
For a single feature layer, they can only retain local structural information.
\section{Structure-Preserving Activation}
\label{sec:method}

\subsection{Overview}
On the basis of the structure-preserving ability of convolutional features, we obtain precise localization maps by proposing the SPA approach for WSOL. 
As shown in Fig.~\ref{fig:overview}, we adopt the CAM network~\cite{zhou2016learning} as our baseline and remove the last fully-connected layer following ACoL~\cite{zhang2018adversarial}. 
Overall, the proposed SPA retains the structural information of objects in two stages. 
First, as shown in the training phase of Fig~.\ref{fig:overview}, we design the RAM to alleviate the structure-missing issue of the head structure of CAM. 
Furthermore, we propose a restricted activation loss ($L_{RA}$) to cooperate with cross-entropy loss ($L_{CE}$) for driving the model to cover object extent during training phase. 
The total loss of SPA training is defined as:
\begin{equation}
    L = L_{CE} + \alpha L_{RA},
\end{equation}
where $L_{CE}$ is the multi-class cross entropy loss, and $L_{RA}$ is the restricted activation loss in RAM. 
$\alpha$ is a regularization factor to balance the two items. 
Second, as shown in the inference phase of Fig.~\ref{fig:overview}, we propose the self-correlation map generating module (SCG) to obtain accurate localization maps on the basis of the results of CAM during inference phase. 
We extract first- and second-order self-correlation for each point of CAM from the convolutional features and aggregate them to acquire activation maps for object localization.

\subsection {Restricted Activation Module}
\label{sec:ral}
The proposed RAM alleviates the structure-missing issue of CAM from two aspects: suppressing the response value range of CAM to avoid the misleading by the local extremely high response for the classification; discriminating the object from background region with the help of coarse pseudo-masks. 

Given a fully convolutional network (FCN), we denote the last convolutional feature maps as $F$ $\in \mathbb{R}^{H \times W \times C}$, where $H \times W$ is the spatial size, and $C$ is the number of channels which is equal to the number of target classes.
We feed the feature maps into a GAP~\cite{lin2013network} layer followed by a \emph{softmax} layer for classification, as shown in Fig.~\ref{fig:overview}.
To ensure that the high activation value after GAP layer is due to broad object activation rather than the local extremely high response, we first suppress the feature value range using the \emph{sigmoid} layer:
\begin{equation}
    F^{'}_t = sigmoid(F_t),
\end{equation}
where $t$ is the ground truth label index and $F_t$ is the $t_{th}$ feature map.
The \emph{sigmoid} layer can effectively suppress the extremely high response and normalize the values to (0,1).
The GAP layer does not separate the representation of the context from the object~\cite{wang2020robust}, which hinders the model differentiating the object from background. 
To overcome this issue, we propose a simple method to generate coarse pseudo-masks for guiding the model to focus on the object regions. 
The mask generation method is based on the observation that the activation values within the background area are distributed much evenly across all classes, and the activation value of the object region can always be highly responsive in at least one target class.
Therefore, we obtain the coarse background mask $M_{bg}$ as:
\begin{equation}
    M_{bg} = \mathbb{I}(Var(softmax(F))<\tau),
\end{equation}
where $\mathbb{I}(\cdot)$ is the indicator function, and $Var(\cdot)$ denotes the standard deviation of each position on the feature map in the channel dimension. $\tau$ is a constant value as the threshold to determinate the background region.
We further obtain the coarse object region as:
\begin{equation}
    M_{obj} = \mathbb{I}(Var(softmax(F))>\tau+\sigma),
\end{equation}
where $\sigma$ denotes the gap between the background and object regions.
On the basis of pseudo-masks, we define the restricted activation loss to guide the model to focus on the object region as:
\begin{equation}
    L_{RA} = \frac{1}{HW}\sum_{i, j}(M_{bg} * F^{'}_t + M_{obj} * (1- F^{'}_t))|_{i,j},
\end{equation}
where $*$ indicates element-wise multiplication.
The simple $L_1$ regularization loss can guide the model to suppress the background area while paying much attention to active full object extent.
Cooperating with the classification branch, the proposed RAM enables the model to preserve the structural information of the target object.

\subsection{Self-correlation Map Generating}
\label{sec:scm}
\begin{figure}[!t]
\centering
\includegraphics[width=0.99\linewidth]{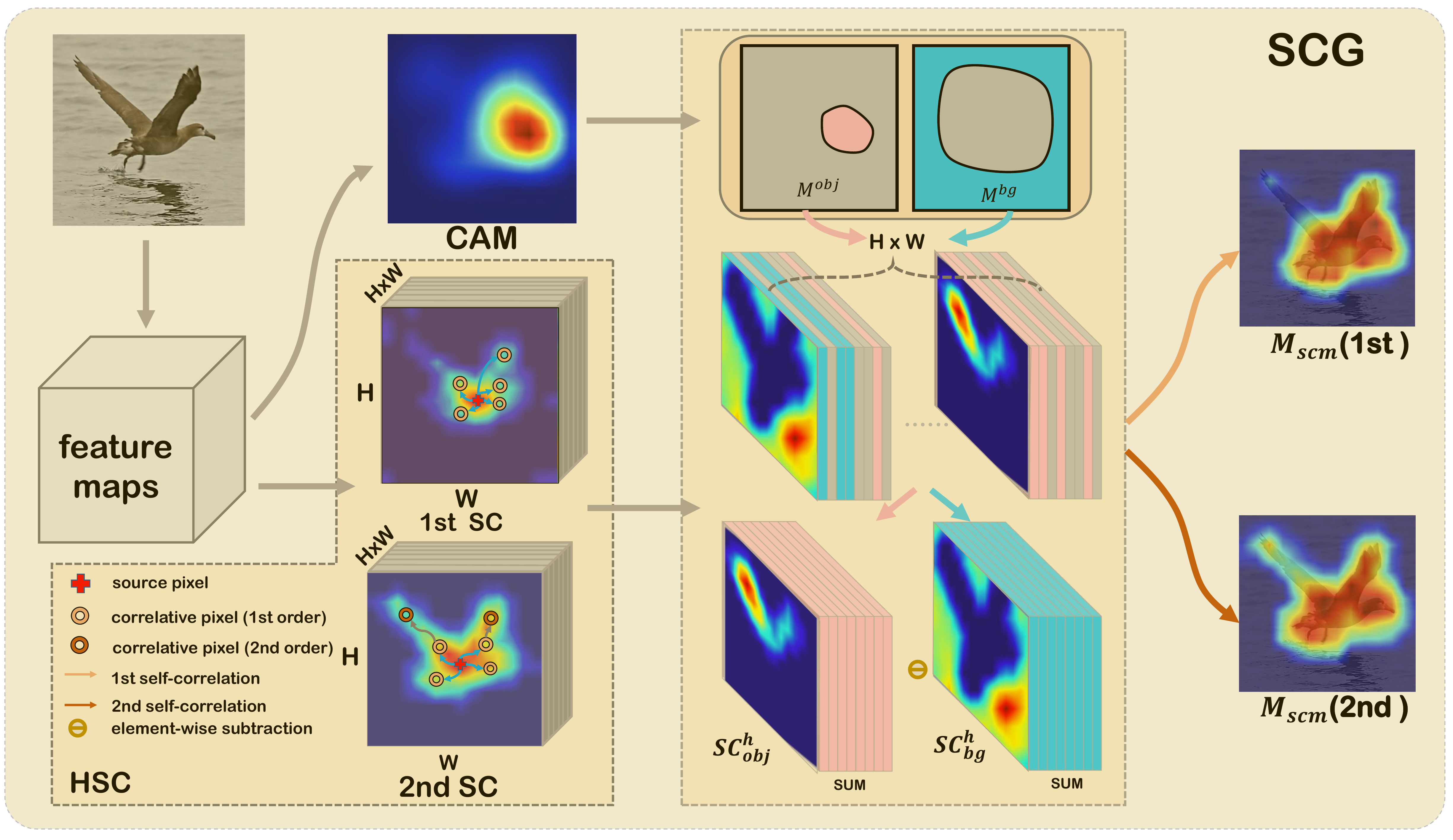}
\caption{Pipeline of the proposed SCG module. Here we show examples of using first- and second-order SC to obtain final localization maps, respectively.}
\label{fig:scm}
\end{figure}
Before introducing SCG, we first analyze the first-order self-correlation($SC^1$) and introduce the concept of spatial HSC, which could capture the long-range structural information on the basis of rich context of the object.
Then, for generating precise localization maps on the basis of HSC, we propose SCG to distill the structure-preserving ability of deep CNN.
Given that we mainly utilize first- and second-order SC in our experiments, we here depict $SC^2$(refer to supplementary for the definition of general high-order $SC$).
\paragraph{First-order Self-correlation.} 
We refer the relation response directly calculated by pixel-to-pixel similarity as spatial \emph{first-order self-correlation} ($SC^1$).
Given a feature map $f \in \mathbb{R}^{HW \times C}$, we use cosine distance to evaluate inter-pixel similarity for feature of index $i$ and $j$:
\begin{equation}
    S(f_i, f_j) = \frac{{f_i}^\mathrm{T}f_j}{||(f_i)||\cdot||(f_j)||},
\end{equation}
where $i,j \in \{0,1,\ldots, HW-1 \}$ indicate the index of features, and $f_i, f_j \in \mathbb{R}^{C \times 1}$ are the feature vectors. 
We define the first-order self-correlation of $f$ as:
\begin{equation}
\begin{aligned}
    &SC^1(f) = \left [ SC^1(f)_{i,j} \right ], \\
where  ~&~ SC^1(f)_{i,j} = ReLU(S(f_i, f_j)) .
\end{aligned}
\label{equ:sc1}
\end{equation} 

The similarities $S(\cdot,\cdot)$ are activated by ReLU~\cite{nair2010rectified} to suppress negative values and $SC^1(f) \in \mathbb{R}^{HW \times HW}$.
Given the large appearance variation, the pixels within an object are usually dissimilar. 
The \engordnumber{2} row in Fig.~\ref{fig:hsc_intro} shows several examples of self-correlation corresponded to positions masked by the red cross. 
It shows that first-order self-correlation can only preserve local spatial structure information.
\paragraph{Second-order Self-correlation.}
To apply the inherent structure-preserving ability of the network for accurate WSOL, we propose to use second-order self-correlation ($SC^2$) to capture long-range structural information of objects. 
 The second order similarity between $f_i$ and $f_i$ are formulated as:
\begin{small}
\begin{equation}
    S^{2}(f_i, f_j) = \frac{1}{(HW)} \sum_{k\in \Omega}S(f_i, f_{k}) \cdot S(f_{k},f_j),
\end{equation}
\end{small}
where $i\neq k \neq j$ and $\Omega$ denotes the set of indexes of all features. 
The $S^{2}(f_i, f_j)$ is then normalized to $[0, 1]$ following:
\begin{equation}
    \hat{S}^{2}(f_i, f_j) = \frac{S^{2}(f_i, f_j) - min_{k \in \Omega}S^{2}(f_i, f_k) }{max_{k \in \Omega}S^{2}(f_i, f_k) - min_{k \in \Omega}S^{2}(f_i, f_k)},
\end{equation}
Then, we define $SC^2$ as:
\begin{equation}
    {SC}^2(f) = \left [ \hat{S}^2(f_i, f_j)) | _{i,j} \right ].
    \label{equ:sc2}
\end{equation}
The \engordnumber{3} row in Fig.~\ref{fig:hsc_intro} lists numerous examples of $SC^2$.
Compared with $SC^1$, $SC^2$ can preserve the details of the object by considering long-range context. 
However, the $SC^2$ may introduce additional noise.
Therefore, we utilize $SC^1$ and $SC^2$ by combining them using element-wise maximum operation in our experiments.

CAM~\cite{zhou2016learning} can only highlight the local region of interest and thus lose the structural information.
To acquire accurate object extent, we propose the \emph{SCG} to refine the localization maps with the help of $HSC$ which is defined as:
\begin{equation}
    HSC_{i,j} = max(SC^1_{i,j}, SC^2_{i,j}),
\end{equation}
where $SC^1$ and $SC^2$ are defined by Eqs.~\ref{equ:sc1} and ~\ref{equ:sc2}.
For clarity of the description, we here reshape ${HSC}(f)$ to $\mathbb{R}^{H \times W \times H \times W}$.
We first employ the CAM to obtain the coarse localization map $M_{cam} \in \mathbb{R}^{H \times W}$ following ACoL~\cite{zhang2018adversarial} by removing the last fully-connected layer.
We define a threshold $\delta_{h}$  to discover the coarse object mask $M^{obj}_{cam} = M_{cam}> \delta_h$. 
Given the indices of object region, we extract the corresponded  $HSC$ of the object as:
\begin{equation}
    HSC_{obj} = G(HSC, M^{obj}_{cam}),
\end{equation}
where $HSC_{obj} \in \mathbb{R}^{N \times H \times W}$. 
$G(\cdot)$ denotes the index function, and $N$ is the number of pixels within object region.
Then the self-correlation map of the object $M_{scg}^{obj}$ is obtained by aggregating the $HSC$ of each point within object region as:
\begin{equation}
    M_{scg}^{obj} = \frac{1}{N}\sum_{i} HSC_{obj}[i].
\end{equation}
To remove the possible background area covered by $M_{scg}^{obj}$, we define another threshold $\delta_l$ and obtain the background self-correlation map $M_{scg}^{bg}$ in a similar way.
We acquire the final localization map $M_{scg}$ as:
\begin{equation}
    M_{scg} = ReLU(M_{scg}^{obj} - M_{scg}^{bg})
\end{equation}
The final self-correlation map $M_{scg}$ is refined by removing the background area and is activated using ReLU to suppress negative values.
Algorithm~\ref{alg:scg_alg} illustrates the procedure of the proposed \emph{SCG} approach.

\begin{algorithm}[htb]  
  \caption{ Localization algorithm of SCG.}  
  \label{alg:scg_alg}  
  \begin{algorithmic}[1]  
    \Require  
      Coarse localization map $M_{cam} \in \mathbb{R}^{H \times W }$;  
      feature map $f \in  \mathbb{R}^{H \times W \times C }$;
      threshold $\delta_{h}$ and $\delta_{l}$; 
    \Ensure  
      Final localization map $M_{scg}$;  
    \State Obtain high-order self-correlation $HSC \in \mathbb{R}^{HW \times HW }$
    \State Reshape $HSC \in \mathbb{R}^{H \times W \times H \times W } \gets reshape(HSC)$
    \State Discover the coarse object region $M^{obj}_{cam} \gets M_{cam} > \delta_{h}$  
    \State Extract object HSC $HSC_{obj} \gets G(HSC,M^{obj}_{cam})$
    \State Obtain the object map $M_{scg}^{obj} \gets sum(HSC_{obj})$
    \State Discover background region $M^{bg}_{cam} \gets M_{cam} < \delta_{l}$
    \State Extract background HSC $HSC_{bg} \gets G(HSC,M^{bg}_{cam})$
    \State Obtain the background map $M_{scg}^{bg} \gets  sum(HSC_{bg}) $
    \State Obtain localization map $M_{scg} \gets (M_{scg}^{obj}-M_{scg}^{bg})_{(>0)}$ 
    \Return $M_{scg}$; 
  \end{algorithmic}  
\end{algorithm}

\section{Experiments}
\label{sec:exp}
\begin{figure}[!tb]
\centering
\includegraphics[width=0.92\linewidth]{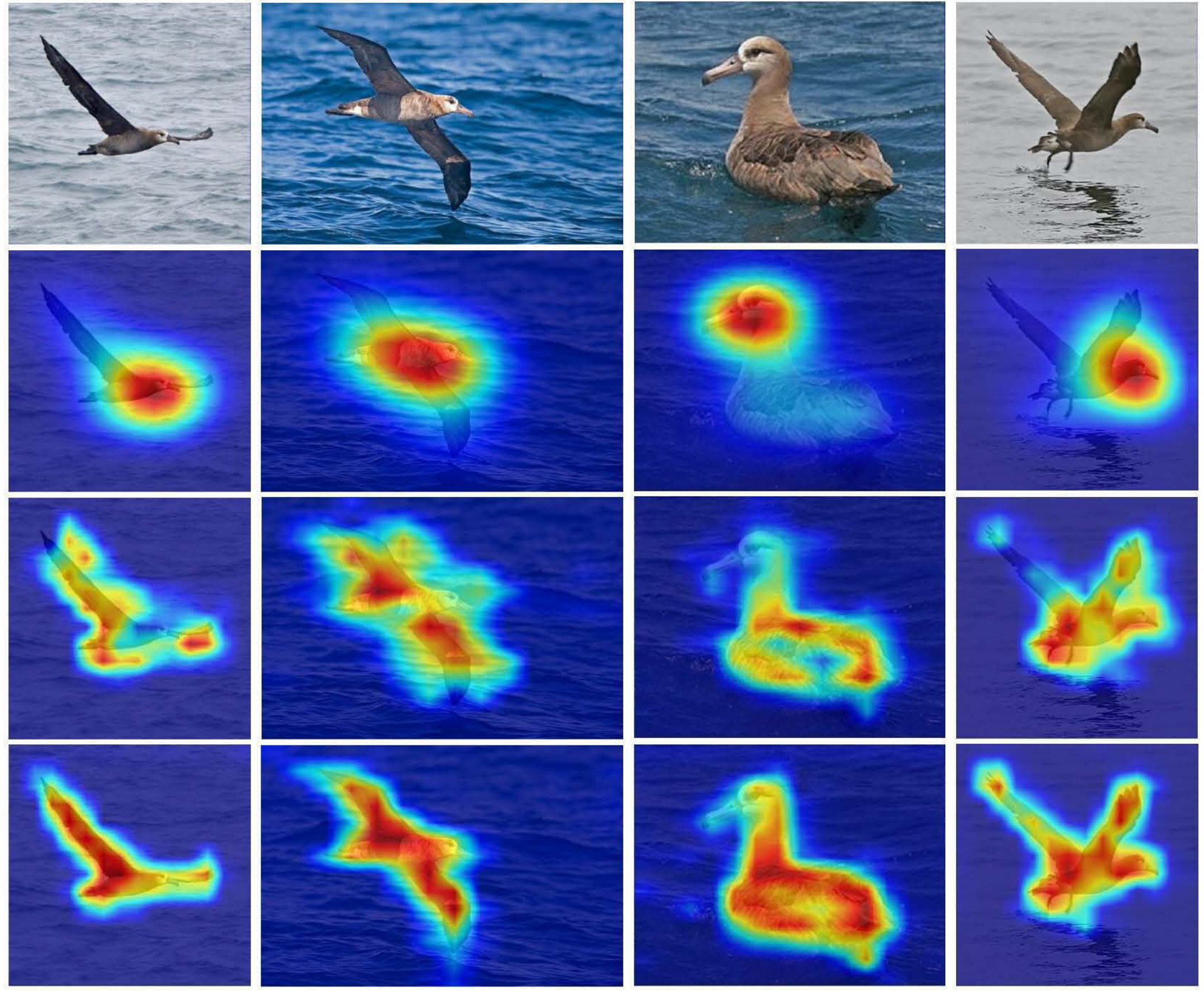}
\caption{Visualization of the localization maps of CAM~\cite{zhou2016learning}, SPA with first- and second-order self-correlation, respectively. The images are from the CUB-200-2011~\cite{wah2011caltech} testing set. }
\label{fig:scm_cub}
\end{figure}
\subsection{Experimental Settings}

\label{sec:exp_setup}
\noindent\textbf{Datasets.} 
We evaluate the proposed approach on two publicly available benchmarks including CUB-200-2011~\cite{wah2011caltech} and ILSVRC~\cite{russakovsky2015imagenet}, following the previous SOTAs~\cite{choe2020evaluating,mai2020erasing,zhang2020inter,zhang2020rethinking}.
CUB-200-2011 is a fine-grained bird dataset of $200$ different species, which is split into the training set of $5,994$ images and the testing set of $5,794$ images.
For ILSVRC, there exist around $1.2$ million images of $1,000$ categories for training and $50,000$ images for validation.
Both benchmarks are all only annotated with class labels for training.
In addition to class labels, CUB-2000-2011 provides the tight box and mask labels for images in testing set.
For ILSVRC, only tight box labels are provided for validation.
Zhang \etal~\cite{zhang2020rethinking} annotated the ground-truth masks for the images on the validation set of ILSVRC: $5,729$ images are manually excluded and the rest are split into the validation ($23,151$ images) and testing sets ($21,120$ images).

\noindent\textbf{Metrics.}
We apply two kinds of metrics to evaluate the localization maps from the bounding box and mask, respectively.
For bounding boxes, we follow the baseline methods~\cite{russakovsky2015imagenet,zhang2018self,zhou2016learning} and report the location error (Loc. Err.).
A prediction is positive when it satisfies the following two conditions simultaneously: the predicted classification labels match the ground-truth categories; the predicted bounding boxes have over 50$\%$ IoU with at least one of the ground-truth boxes. 
$Gt$-$Known$ indicates it considers localization only regardless of classification.
For masks, we mainly utilize Peak-IoU and Peak-T, which are defined in SEM~\cite{zhang2020rethinking}, to directly evaluate the localization performance by performing a pixel-wise comparison between the predicted localization map and the ground-truth mask. 
\emph{Peak-IoU} $\in [0, 1]$ and \emph{Peak-T} $\in [0, 255]$ denote the best IoU score and its corresponding threshold, respectively.
A high-quality localization map should meet two requirements: 1) the full object extent can be accurately covered with a specific threshold; 2) brightness values of pixels belonging to object and the background should differ greatly so that the objects can be well visualized~\cite{zhang2020rethinking}.
High \emph{Peak-IoU} and \emph{Peak-T} values indicates good localization maps. 
\begin{table}[!t]
\centering
\resizebox{.46\textwidth}{!}{
\begin{tabular}{l|c|c|c|c}
	\toprule
      \multirow{2}{*}{Methods} &\multirow{2}{*}{Backbone}  &\multicolumn{3}{|c}{Loc Err.}  \\ 
      \cline{3-5}
       &  &Top-1 & Top-5&Gt-Known  \\ \hline
       Backprop~\cite{simonyan2013deep} & VGG16 & 61.12 & 51.46 & -  \\
       CAM~\cite{zhou2016learning} & VGG16 &57.20 &45.14 &-  \\
       CutMix~\cite{yun2019cutmix} & VGG16 &56.55 &- & - \\
       ADL~\cite{choe2019attention} & VGG16 & 55.08 & -&- \\
       ACoL~\cite{zhang2018adversarial} &VGG16 &54.17 &40.57 & 37.04  \\
       I$^{2}$C~\cite{zhang2020inter} & VGG16 &52.59 &41.49 & 36.10  \\
       MEIL~\cite{mai2020erasing} &VGG16 &53.19 & - & -  \\
       Ours & VGG16 &{\bf 50.44} &{\bf 38.68} & {\bf 34.95}  \\ \hline
       CAM~\cite{zhou2016learning} &InceptionV3 &53.71 & 41.81 &37.32 \\
       SPG~\cite{zhang2018self} & InceptionV3 &51.40 & 40.00 & 35.31  \\
       ADL~\cite{choe2019attention} &InceptionV3 &51.29 & - & -  \\
       ACoL~\cite{zhang2018adversarial} & GoogLeNet & 53.28 & 42.58 & -  \\
       DANet~\cite{xue2019danet} &GoogLeNet & 52.47 & 41.72 &-  \\
       MEIL~\cite{mai2020erasing} &InceptionV3 & 50.52 & - &-  \\
       I$^{2}$C~\cite{zhang2020inter} & InceptionV3 & {\bf 46.89} &35.87 &{\bf 31.50}  \\
       GC-Net~\cite{lu2020geometry} & InceptionV3 & 50.94 & 41.91 & -  \\
       Ours & InceptionV3 & 47.27 & {\bf 35.73} & 31.67 \\
	\bottomrule
\end{tabular}
}
\caption{Comparison between our method and the state-of-the-art on the ILSVRC~\cite{russakovsky2015imagenet} validation set.
}
\label{tab:ilsvrc_main}
\end{table}

\noindent\textbf{Implementation Details.}
We implement the proposed algorithm on the basis of two popular backbone networks, $i.e.$, VGG16~\cite{simonyan2014very} and Inception V3~\cite{szegedy2016rethinking}. 
We make the same modifications on backbones following ACoL~\cite{zhang2018adversarial} and SPG~\cite{szegedy2016rethinking}, and use the simplified method in ACoL~\cite{zhang2018adversarial} to obtain localization maps.
Both networks are fine-tuned on the pre-trained weights of ILSVRC~\cite{russakovsky2015imagenet}. 
The input images are randomly cropped to $224 \times 224$ pixels after being re-sized to $256 \times 256$ pixels. 
For classification, we average the scores from the softmax layer with $10$ crops.
We also implement several recent benchmark methods, $i.e.$, CAM~\cite{zhou2016learning}, HaS~\cite{singh2017hide}, ACoL~\cite{zhang2018adversarial}, SPG~\cite{szegedy2016rethinking}, ADL~\cite{choe2019attention}, and CutMix~\cite{yun2019cutmix} in accordance with the codes\footnote{\url{https://github.com/clovaai/wsolevaluation}} released by Choe \etal~\cite{choe2020evaluating}. 
For fair comparisons, we adopt the same training strategy with SEM~\cite{zhang2020rethinking}.
The codes for \emph{Peak-IoU} and \emph{Peak-T} are provided on the workshop of \emph{Learning from Imperfect Data (LID)}\footnote{\url{https://lidchallenge.github.io/challenge.html}}.   
To calculate the self-correlation on VGG16, we utilize the features of \textit{Stages $4$} and \textit{$5$}, and combine the two $HSC$s by element-wise summation.
For Inception V3, we utilize the features of layer \textit{feat4} and \textit{feat5} to calculate HSC and sum them element wise. 

\begin{figure*}[!thb]
\centering
\includegraphics[width=0.92\linewidth]{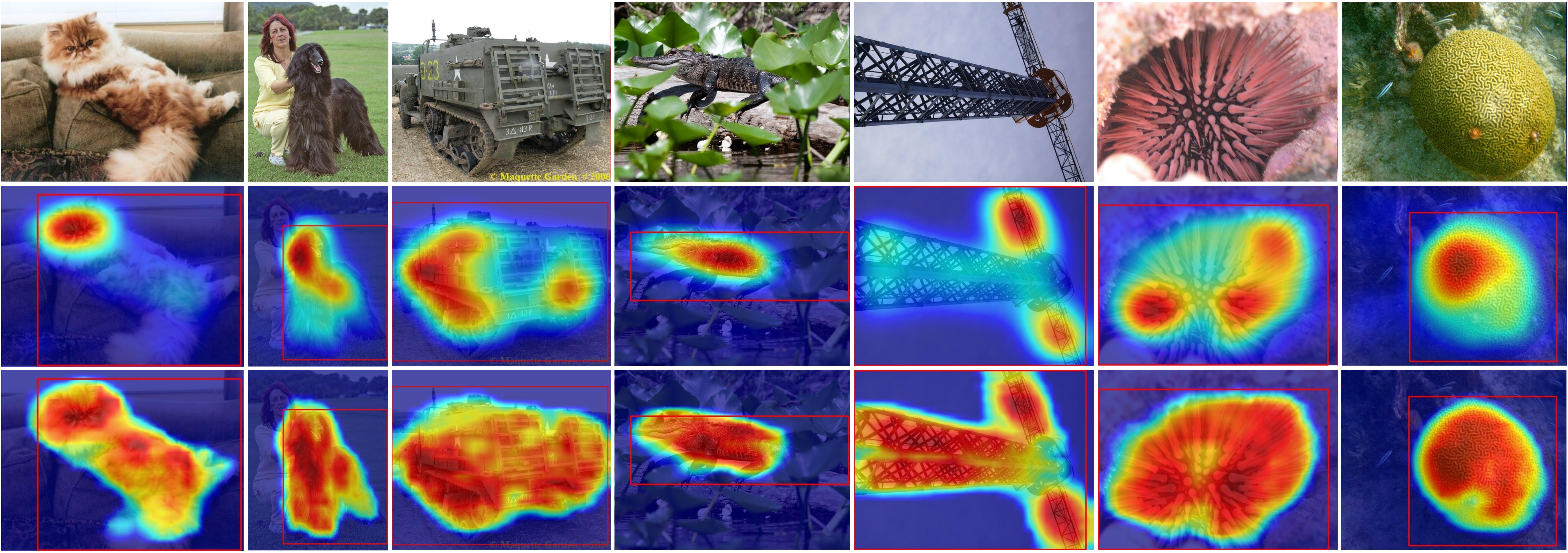}
\caption{Visualization of the localization maps with CAM~\cite{zhou2016learning} (middle row) and the proposed SPA (bottom row). The ground truth boxes are in red. The images are from the ILSVRC~\cite{russakovsky2015imagenet} validation set.}
\label{fig:scm_ilsvrc}
\end{figure*}
\subsection{Experimental Results}
\begin{table}
\centering
\resizebox{.46\textwidth}{!}{
\begin{tabular}{l|c|c|c|c}
	\toprule
      \multirow{2}{*}{Methods} &\multirow{2}{*}{Backbone}  &\multicolumn{3}{|c}{Loc Err.}   \\ 
      \cline{3-5}
       &  &Top-1 & Top-5&Gt-Known \\ \hline
       CAM~\cite{zhou2016learning} & GoogLeNet & 58.94 & 49.34 &44.9  \\
       SPG~\cite{zhang2018self} & GoogLeNet & 53.36 &42.8 & -  \\
       DANet~\cite{xue2019danet} & InceptionV3 & 50.55 &39.54 & 33.0  \\
       ADL~\cite{choe2019attention} & InceptionV3 & 46.96 & - & -  \\ 
       Ours & InceptionV3 & {\bf 46.41} &{\bf 33.50} &{\bf 27.86}   \\ \hline
       CAM~\cite{zhou2016learning} & VGG16 &55.85 &47.84 &44.0  \\
       ADL~\cite{choe2019attention} & VGG16 & 47.64 & -&- \\
       ACoL~\cite{zhang2018adversarial} &VGG16 &54.08 &43.49 & 45.9 \\
       DANet~\cite{xue2019danet} &VGG16 &47.48 &38.04 & 32.3  \\
       SPG~\cite{zhang2018self} & VGG16 & 51.07 &42.15 & 41.1  \\
       I$^{2}$C~\cite{zhang2020inter} & VGG16 &44.01 &31.6 & -  \\
       MEIL~\cite{mai2020erasing} &VGG16 &42.54 & - & -  \\
       GC-Net~\cite{lu2020geometry} &VGG16 & {\bf 36.76} & {\bf 24.46} &{\bf 18.9}  \\
       Ours & VGG16 & 39.73 & 27.5 &  22.71 \\
	\bottomrule
\end{tabular}}
\caption{Comparison between our method and the state-of-the-art on the CUB-200-2011~\cite{wah2011caltech} test set.
}
\label{tab:cub_main}
\end{table}

\noindent\textbf{Bounding Box Localization.}
We first compare the proposed approach with the SOTAs on the localization error by using tight bounding boxes. 
We only show the \emph{Loc. Err.} (refer to the supplementary materials for more details). 
Table~\ref{tab:ilsvrc_main} reports the results of our method and several baselines on the ILSVRC validation set. 
Our method, on the basis of VGG16, achieves the lowest error rate of $50.44\%$ in Top-1 \emph{Loc. Err.}, significantly surpassing all the baselines. 
Specifically, we achieve remarkable gains of $3.5\%$ and $4.4\%$ in terms of Top-1 \emph{Loc. Err.} compared with ACoL and ADL. 
Compared with the state-of-the-art $I^2C$, we achieve a performance gain of $2.0\%$, which is a significant margin to the challenging problem.
On the InceptionV3, our method obtains comparable results with $I^2C$ and surpasses other methods significantly. 
$I^{2}C$ leverages pixel-level similarities across different objects to prompt the consistency of object features within the same categories, but it cannot retain the structural information for the objects. 
Fig.~\ref{fig:scm_ilsvrc} shows several examples of the localization maps by CAM~\cite{zhou2016learning} and the proposed SPA. 
Our results retain the structure of objects well and cover more extent of the objects. 

Table~\ref{tab:cub_main} compares the proposed method with various baseline methods on the CUB-200-2011 testing set. 
All the baselines adopt CAM to obtain localization maps.
Our method, on the basis of VGG16, surpasses all the baseline methods on Top-1, Top-5, and Gt-Known metrics, yielding the localization error of Top-1 $39.73\%$, and Top-5 $27.5\%$.
Compared with the current state-of-the-art $I^{2}C$ and $MEIL$, we achieve gains of 3.5$\%$ and 2.0$\%$ in terms of Top-1 \emph{Loc. Err.}, respectively. 
Fig.~\ref{fig:scm_cub} shows some examples of the localization map. 
The \engordnumber{3} and \engordnumber{4} rows are the results of our method by using first- and second-order self-correlation, respectively.
Compared with CAM~\cite{zhou2016learning}, the results of our method preserve the structure of objects well. 
The results of $SC^2$ obtain more accurate masks than that of $SC^1$, but they
obtain almost the same tight bounding boxes.
To reveal the superiority of the method, we further evaluate our method by comparing with the ground-truth masks below.

\begin{figure}[!tb]
\centering
\subfigure[Image]{
\includegraphics[width=0.17\linewidth]{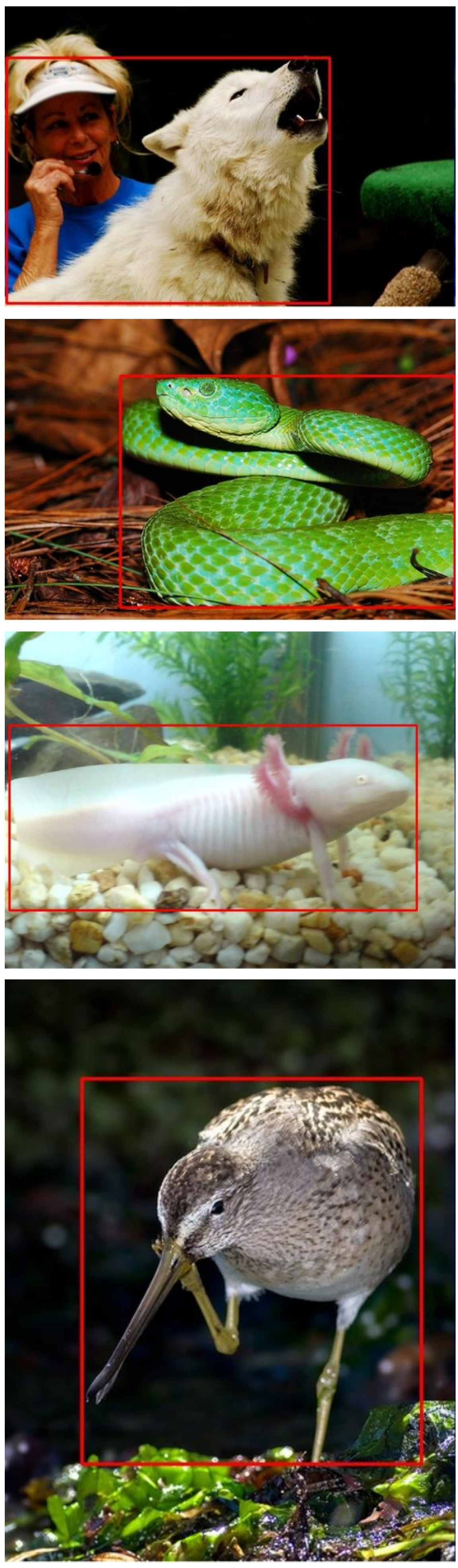}
\label{fig:scm_pro_img}}
\subfigure[CAM]{
\includegraphics[width=0.17\linewidth]{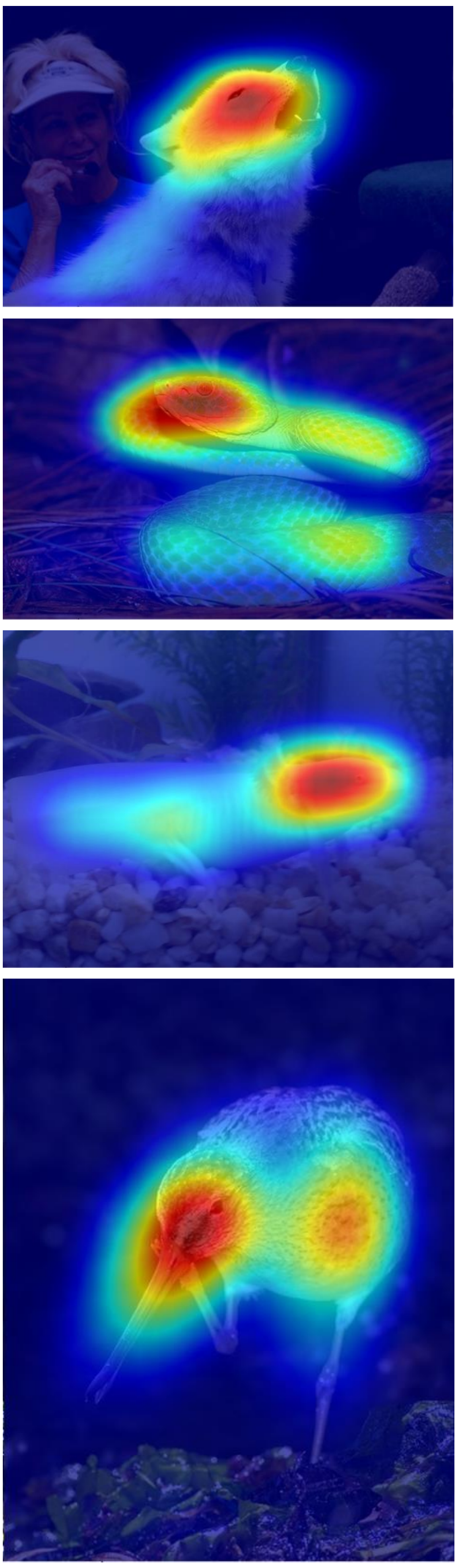}
\label{fig:scm_pro_cam}}
\subfigure[$M_{scm}^{obj}$]{
\includegraphics[width=0.17\linewidth]{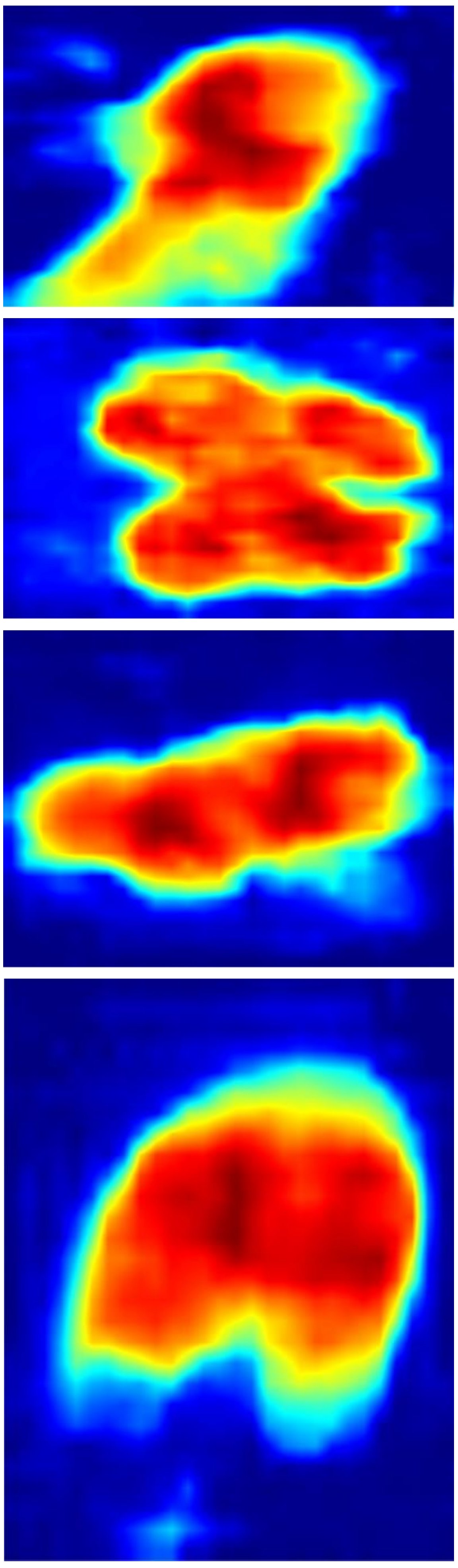}
\label{fig:scm_pro_obj}}
\subfigure[$M_{scm}^{bg}$]{
\includegraphics[width=0.17\linewidth]{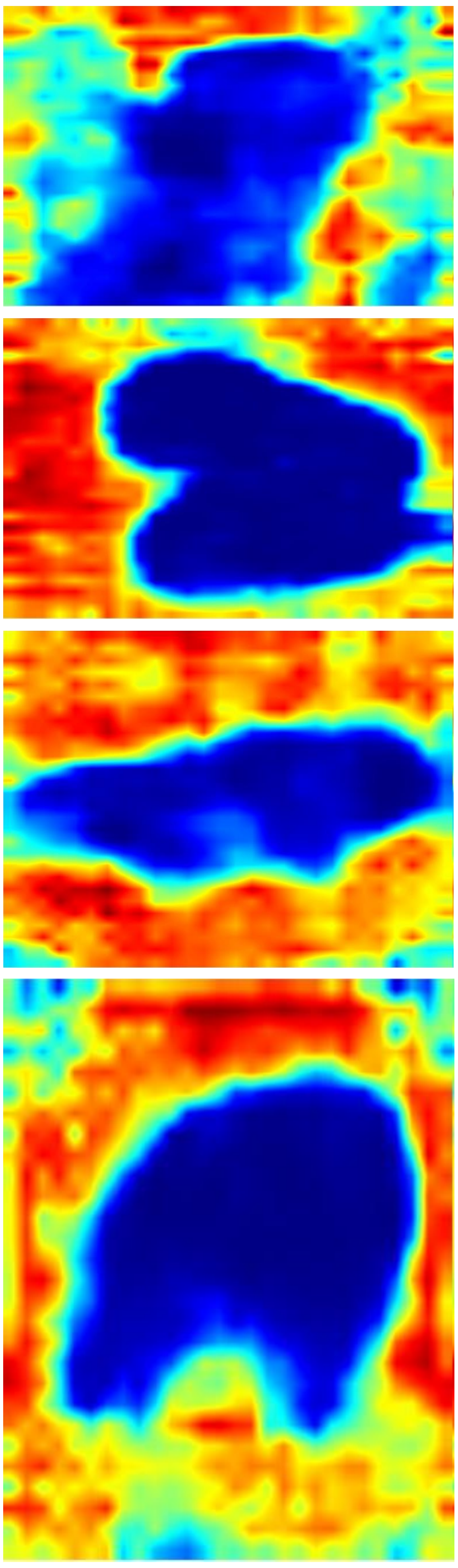}
\label{fig:scm_pro_bg}}
\subfigure[$M_{scm}$]{
\includegraphics[width=0.17\linewidth]{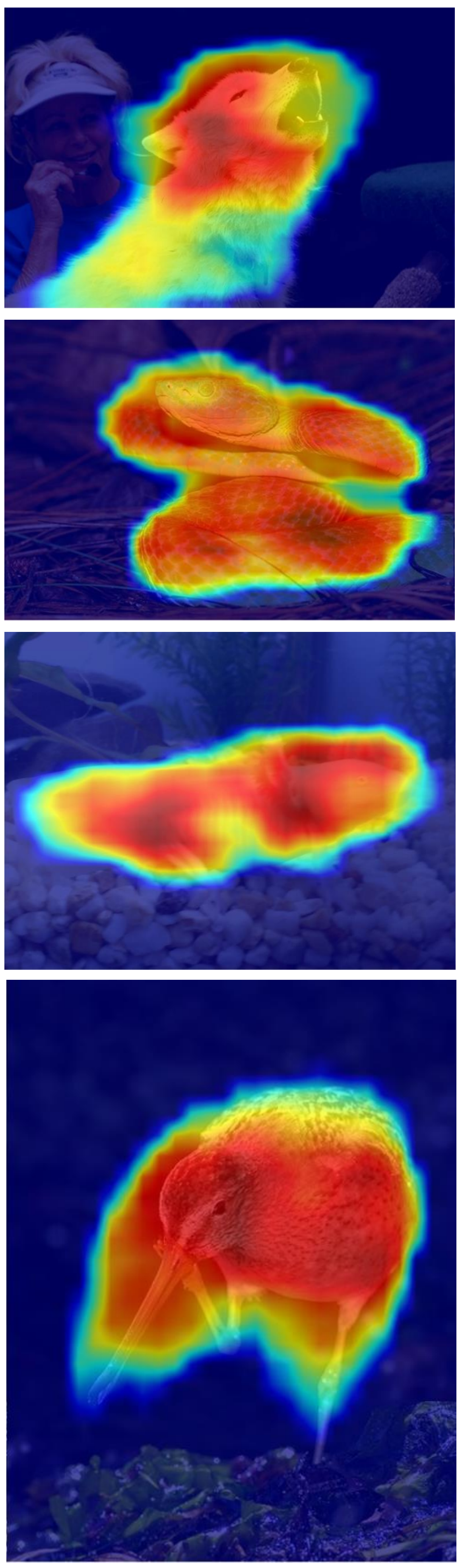}
\label{fig:scm_pro_scm}}
\caption{Visualization of process for SCG. Given the input images (a), we employ simplified CAM~\cite{zhou2016learning} to obtain the coarse localization maps (b). (c) and (d) are the object and background SC maps by aggregating HSCs of pixels within the corresponding areas on the basis of CAM, respectively. 
The final SCGs (e) are obtained by subtracting background from object SC maps.}
\label{fig:scm_process}
\end{figure}
\begin{table}
\centering
\resizebox{.45\textwidth}{!}{
\begin{tabular}{lcc|ccc}
	\toprule
     Methods &SEM~\cite{zhang2020rethinking} &SCG  & Peak-IoU & Peak-T &GT-Known \\     \hline
     CAM~\cite{zhou2016learning} & &  & 53.59 & 33 & 64.09 \\\hline
     & $\checkmark$ & & 51.39 & 74 & 62.67 \\
     & &$\checkmark$  & {\bf56.38} & {\bf79} & {\bf66.79}\\ \hline
     HaS~\cite{singh2017hide} & &  &54.99 &50 &65.32  \\ \hline
     & $\checkmark$ && 51.59 & 79 & 63.06     \\
     & &$\checkmark$ & {\bf57.29} & {\bf92} & {\bf68.31}  \\ \hline
     ACoL~\cite{zhang2018adversarial} & &  &50.89 & 52 & 63.28 \\ \hline
     & $\checkmark$ & & 48.92 & 83 & 59.92    \\
     & &$\checkmark$ & {\bf52.45} & {\bf132} & {\bf65.45}  \\ \hline
     CutMix~\cite{yun2019cutmix} & &  & 54.54 & 34 & 64.65 \\ \hline
     & $\checkmark$ & & 52.02  & 79 & 63.84    \\
     & &$\checkmark$ &{\bf56.96} & {\bf83}  & {\bf68.23}  \\ \hline
     SPG~\cite{zhang2018self} & &  &53.76 & 33 & 64.19 \\ \hline
     & $\checkmark$ & & 51.74 & 84 & 63.08    \\
     & &$\checkmark$ &{\bf56.40} & {\bf89}  & {\bf66.78} \\ \hline
     ADL~\cite{choe2019attention} & & & 52.87 & 29 & 63.64 \\ \hline
     & $\checkmark$ & & 50.01 & 72 & 62.38   \\
     & &$\checkmark$ & {\bf56.01}  & {\bf76} & {\bf66.08} \\
	\bottomrule
\end{tabular}
}
\caption{Evaluation results of Peak-T, Peak-IoU and GT-Known \emph{Loc Acc} on ILSVRC validation set. All the methods apply Inception V3 as the backbone network. }
\label{tab:ilsvrc_mask}
\end{table}
\noindent\textbf{Mask Localization.}
To further verify the effectiveness of our method, we compare the localization map with the ground-truth mask and adopt the \emph{Peak-T} and \emph{Peak-IoU} as metrics following SEM~\cite{zhang2020rethinking}. 
We also report the \emph{Gt-Known Loc. Acc.} of each method.
In this section, we only apply the SCG to the baseline methods without involving RAM for fair comparison with SEM.
Given the input images, we employ the simplified CAM following ACoL~\cite{zhang2018adversarial} to obtain the coarse localization maps.
Fig.~\ref{fig:scm_process} visualizes the detailed process of the proposed SCG. 
All baseline methods adopt our re-implemented models and surpass the corresponding results of SEM as shown in Table~\ref{tab:ilsvrc_mask}. 
The proposed SCG achieves consistent gains on \emph{Peak-IoU}, \emph{Peak-T}, and \emph{Gt-Known}.
Specifically, we achieve an improvement by $2.3\%$ compared with the best baseline HaS~\cite{singh2017hide} in terms of \emph{Peak-IoU}.
As for \emph{Peak-T}, our results significantly outperform all the baselines by improving about $50$ points on average.
The re-implemented SEM on the basis of the code released by the author performs worse than all the baselines.

\begin{table}
\centering
\resizebox{.46\textwidth}{!}{
\begin{tabular}{l|ccc|ccc}
	\toprule
    \multirow{2}{*}{Methods} &\multicolumn{3}{|c}{ILSVRC($\%$)}  &\multicolumn{3}{|c}{CUB-2011-200($\%$)} \\
     &M-Ins & Part & More &M-Ins & Part &More\\
    \hline
    VGG16 &10.65 &3.85 &9.58 &- &21.91 &10.53 \\
     Ours & 9.97& 2.83 &7.66 &-  & 9.25 &6.33 \\
    \hline
    InceptionV3 &10.36 &3.22 &9.49 &- &23.09&5.52 \\
     Ours &9.48  & 2.89 &7.80 &-  &12.81  &6.83  \\
	\bottomrule
\end{tabular}
}
\caption{Localization error statistics.}
\label{tab:err_analysis}
\end{table}
\noindent\textbf{Error Analysis.}
To further reveal the effect of our method, we divide the localization error into five cases: classification (Cls), multi-instance (M-Ins), localization part (Part), localization more (More), and other (OT) errors.
\emph{Part} indicates that the predicted bounding box only cover the parts of object, and IoU is less than a certain threshold.
Contrastingly, \emph{More} indicates that the predicted bounding box is larger than the ground truth bounding box by a large margin.
Each metric calculates the percentage of images belonging to the corresponding error in the validation/testing set.  
Table~\ref{tab:err_analysis} lists localization error statistics of \emph{M-Ins}, \emph{Part}, and \emph{More}. 
Our method effectively reduces the \emph{M-Ins}, \emph{Part}, and \emph{More} errors, which indicates that our localization maps are much accurate.
Refer to supplementary materials for detailed analysis and definitions of each metric. 
\subsection{Ablation Study}
We conduct a series of experiments to verify the effectiveness of the proposed RAM and SCG.
Table~\ref{tab:ilsvrc_abl} shows the results on the ILSVRC validation set with different configurations. 
On VGG16, the RAM and SCG improve the baseline by $2\%$ and $1.4\%$, respectively. 
It achieves a significant gain of $3.1\%$ when we use both modules simultaneously. 
On Inception V3, the two modules also achieve remarkable gains, yielding a localization error of Top-1 47.29$\%$.
In Table~\ref{tab:cub_abl}, we evaluate the performance of the RAM and SCG on CUB-200-2011 testing set. 
The proposed approach achieves significant improvements. 
Specifically, the SCG and RAM obtain gains of $11.5\%$ and $8.1\%$ in terms of Top-1 \emph{Loc. Err.} on VGG16 respectively, and it achieves a remarkable improvement of $17.7\%$ when we employ both modules simultaneously.
On Inception V3, our method also achieves a significant gain of $9.2\%$ Top-1 \emph{Loc. Err.}
The experimental results show that the proposed approach achieves consistent and substantial improvement on different backbones and benchmarks.  
Refer to the supplementary materials for more details.

\begin{figure}[!t]
\centering
\includegraphics[width=0.99\linewidth]{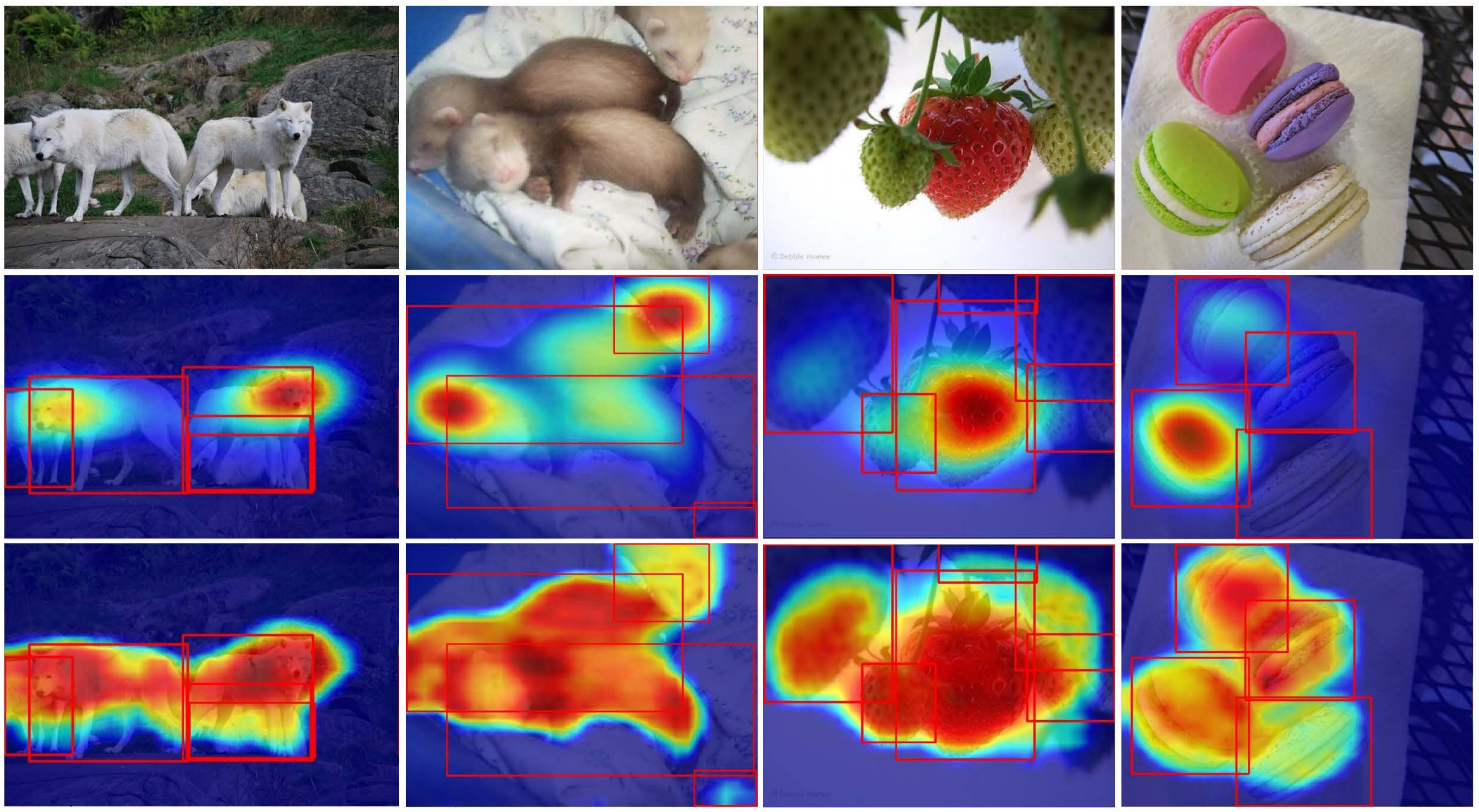}
\caption{Localization maps of some images from ILSVRC validation set with CAM~\cite{zhou2016learning} and the proposed SPA. Ground truth boxes are in red. Both methods cannot separate each instance when multiple instances exist in the scene. }
\label{fig:scm_multi_ins}
\end{figure}

\subsection{Limitation}
Although the proposed approach achieves much better performance than CAM-based SOTAs, it is challenged when multiple instances come together.
Fig.~\ref{fig:scm_multi_ins} shows localization results with CAM and our approach in the multi-instance scenes.
Compared with CAM, our results more precisely cover the object extent. 
However, given the lack of instance-level supervision, distinguishing different instances is difficult.
The results in Table~\ref{tab:err_analysis} also show that the \emph{M-Ins} error is currently the main source of localization error.
The structural information from other images containing only one instance may alleviate this problem. 
In the future work, the consistency of structure preserving across images must be explored to achieve \emph{weakly supervised instance localization}.
\begin{table}
\centering
\resizebox{.46\textwidth}{!}{
\begin{tabular}{lcc|ccc}
	\toprule
      \multirow{2}{*}{Methods} &\multirow{2}{*}{SCG} &\multirow{2}{*}{RAM} &\multicolumn{3}{|c}{Loc Err.}  \\ 
      & & &Top-1 & Top-5 &Gt-Known\\
    \hline
      VGG16 & & &53.76 &42.75 &39.21 \\
    \hline
      & $\checkmark$ & &51.15  &39.57  &35.98  \\
      & &$\checkmark$ & 52.33 &40.88 &37.29 \\
      & $\checkmark$ &$\checkmark$ &{\bf 50.44} &{\bf 38.68} &{\bf 34.95} \\
    \hline
    InceptionV3 & & &49.86 & 38.86 &35.05 \\
    \hline
     & $\checkmark$ & &47.38  &35.75  & 31.74  \\
     & &$\checkmark$ &49.31  & 38.20 & 34.29 \\
     & $\checkmark$ &$\checkmark$ &{\bf 47.27} & {\bf 35.73} & {\bf 31.67} \\
	\bottomrule
\end{tabular}
}
\caption{Localization error on ILSVRC~\cite{russakovsky2015imagenet} validation set when using different configurations.}
\label{tab:ilsvrc_abl}
\end{table}

\begin{table}
\centering
\resizebox{.46\textwidth}{!}{
\begin{tabular}{lcc|ccc}
	\toprule
    \multirow{2}{*}{Methods} &\multirow{2}{*}{SCG} &\multirow{2}{*}{RAM} &\multicolumn{3}{|c}{Loc Err.} \\
     & & &Top-1 & Top-5 &Gt-Known\\
    \hline
    VGG16 & & &57.49 &49.05 &46.24 \\
    \hline
     & $\checkmark$ & &45.98  &35.59  &31.33 \\
     & &$\checkmark$ & 49.34 &39.14 &34.50 \\
     & $\checkmark$ &$\checkmark$ &{\bf 39.73} &{\bf 27.55} &{\bf 22.71}\\
    \hline
    InceptionV3 & & &55.64 &44.27 &39.14 \\
    \hline
      & $\checkmark$ & &48.84  &35.80  & 30.77 \\
      & &$\checkmark$ &52.04  &40.21 &35.17 \\
      & $\checkmark$ &$\checkmark$ &{\bf 46.41} &{\bf 33.50} &{\bf 27.86} \\
	\bottomrule
\end{tabular}
}
\caption{Localization error on CUB-200-2011~\cite{wah2011caltech} test set when using different configurations.}
\label{tab:cub_abl}
\end{table}
\section{Conclusion}
\label{sec:conc}
In this study, we unveiled the fact that the spatial structure-preserving is crucial to discover the localization information contained in convolutional features for WSOL.
We accordingly proposed a structure preserving activation (SPA) approach to precisely localize objects. 
SPA leverages the restricted activation maps to alleviate the structure missing issue of head structure of the classification network. 
It also utilizes self-correlation generation (SCG) to distill the structure-preserving ability of features for acquiring precise localization maps. 
In SCG, second-order correlation is proposed to make up the inability of first-order self-correlation for capturing long-range structural information. 
Extensive experiments on CUB-200-2011 and ILSVRC benchmarks validated the effectiveness of the proposed approach, in striking contrast with the state-of-the-arts. 
The SPA approach provides a fresh insight to the WSOL problem.

\small{
\paragraph{Acknowledgment.}
This work was supported by National Key R\&D Program of China under no. 2018YFC0807500, and by National Natural Science Foundation of China under nos. U20B2070, 61832016, 61832002 and 61720106006, and by CASIA-Tencent Youtu joint research project.
}

{\small
\bibliographystyle{ieee_fullname}
\bibliography{spa}
}

\end{document}